# PyramidMamba: Rethinking Pyramid Feature Fusion with Selective Space State Model for Semantic Segmentation of Remote Sensing Imagery

Libo Wang, Dongxu Li, Sijun Dong, Xiaoliang Meng, Xiaokang Zhang and Danfeng Hong

*Abstract*—Semantic segmentation, as a basic tool for intelligent interpretation of remote sensing images, plays a vital role in many Earth Observation (EO) applications. Nowadays, accurate semantic segmentation of remote sensing images remains a challenge due to the complex spatial-temporal scenes and multi-scale geo-objects. Driven by the wave of deep learning (DL), CNN- and Transformer-based semantic segmentation methods have been explored widely, and these two architectures both revealed the importance of multi-scale feature representation for strengthening semantic information of geo-objects. However, the actual multi-scale feature fusion often comes with the semantic redundancy issue due to homogeneous semantic contents in pyramid features. To handle this issue, we propose a novel Mamba-based segmentation network, namely PyramidMamba. Specifically, we design a plug-and-play decoder, which develops a dense spatial pyramid pooling (DSPP) to encode rich multi-scale semantic features and a pyramid fusion Mamba (PFM) to reduce semantic redundancy in multi-scale feature fusion. Comprehensive ablation experiments illustrate the effectiveness and superiority of the proposed method in enhancing multi-scale feature representation as well as the great potential for real-time semantic segmentation. Moreover, our PyramidMamba yields state-of-the-art performance on three publicly available datasets, i.e. the OpenEarthMap (70.8% mIoU), ISPRS Vaihingen (84.8% mIoU) and Potsdam (88.0% mIoU) datasets. The code will be available at https://github.com/WangLibo1995/GeoSeg.

*Index Terms*—Mamba, State Space Model, Semantic Segmentation, Multi-scale Feature Fusion.

## I. INTRODUCTION

Semantic segmentation of fine-resolution remote sensing images has become increasingly crucial for a wide range of Earth Observation (EO) applications [1], such as land use land cover (LULC) mapping [2], [3], [4], environment monitoring [5], [6], and urban sustainable development [7], [8]. Driven by the evolution of artificial intelligence and sensor technology, deep learning (DL) [9] has been seamlessly integrated into the remote sensing field, acting as a catalyst for the processing and analyzing vast volumes of remote sensing big data [10], [11]. Compared to traditional machine learning methods, such as Support Vector Machines (SVMs) and Random Forests, DL-based approaches demonstrate their superiority in automatic and robust semantic feature extraction, thereby obtaining further accuracy improvements.

The fully convolutional network (FCN) [12] first adopts the DL-based end-to-end manner to construct a novel convolutional neural network (CNN) for semantic segmentation. Since then, FCN-based semantic segmentation methods gradually become mainstream. Although FCN achieved important breakthroughs, its single and limited receptive field results in a coarse segmentation [13]. To address this issue, some studies apply attention mechanisms to introduce global contextual information. Typical methods include the non-local neural network [14], dual attention network [15], and Transformers [16], [17]. Another part of the research employs the pyramid or multi-scale feature fusion scheme to achieve multiple receptive fields. The most representative methods include the pyramid scene parsing network (PSPNet) [18] and the feature pyramid network (FPN) [19]. However, the above methods both have shortcomings. Due to the square complexity of attention mechanisms, the former requires numerous computational resources to model global information, demonstrating lower efficiency. The latter often leads to the redundancy issue in multi-scale feature fusion since there is plenty of homogeneous semantic information in pyramid features. Thus, how to effectively aggregate multi-scale semantic features remains a

This work was supported in part by the National Natural Science Foundation of China (No.41971352) and Alibaba innovation research (AIR:17289315). *(Corresponding author: Xiaoliang Meng).*

Libo Wang and Dongxu Li are with School of Remote Sensing and Geomatics Engineering, Nanjing University of Information Science and Technology, 210044, Nanjing, Jiangsu, China, and also with Technology Innovation Center for Integration Applications in Remote Sensing and Navigation, Ministry of Natural Resources, 210044, Nanjing, Jiangsu, China. (e-mail: rosswanglibo@gmail.com; 202113350064@nuist.edu.cn).

Sijun Dong and Xiaoliang Meng are School of Remote Sensing and Information Engineering, Wuhan University, Wuhan 430079, China. (e-mail: dyzy41@whu.edu.cn; xmeng@whu.edu.cn).

Xiaokang Zhang is with the School of Information Science and Engineering, Wuhan University of Science and Technology, Wuhan 430081, China. (e-mail: natezhangxk@gmail.com).

Danfeng Hong is with the Aerospace Information Research Institute, Chinese Academy of Sciences, Beijing 100094, China, and also with the School of Electronic, Electrical and Communication Engineering, University of Chinese Academy of Sciences, Beijing 100049, China. (email:hongdf@aircas.ac.cn).

challenge.

Recently, a novel architecture based on the selective state space models (SSM) [20], namely Mamba, has attracted widespread attention in the fields of computer vision and natural language processing. Unlike Transformers that apply inefficient self-attention mechanisms for sequence modeling, Mamba takes advantage of the selective scan mechanism, unique hardware-aware algorithm, and parallel scanning, demonstrating immense advantages in processing long sequences with high efficiency. In particular, the selective scan mechanism allows the Mamba to compress homogeneous features and extract core semantic information. Thus, this scheme has great potential in addressing the redundancy issue in multi-scale feature fusion.

In this paper, we construct a Mamba-based network, namely PyramidMamba, for semantic segmentation of remote sensing images. Specifically, we developed a Mamba-based decoder, including a dense spatial pyramid pooling (DSPP) module and a pyramid fusion mamba (PFM) module. The DSPP allows more pooling scales compared to the standard spatial pyramid pooling module, thereby capturing more fine-grained multi-scale contexts. The PFM introduces the standard Mamba block to aggregate pyramid semantic features to relieve the redundancy issue and enhance multi-scale visual representation. Moreover, plug-and-play DSPP and PFM can be integrated into deep neural networks for efficient and effective multi-scale feature representation. The main contributions of this paper can be summarized as follows:

1) We rethink the pyramid feature fusion schemes and developed a novel Mamba-based segmentation network (PyramidMamba) to improve multi-scale feature representation.
2) We design a Mamba-based decoder that applies dense spatial pooling for producing more fine-grained multi-scale contexts while using the selective characteristic of Mamba to reduce homogeneous semantic information effectively in multi-scale feature fusion. Besides, benefiting from the efficient sequence modeling of Mamba, this decoder also demonstrates great potential in building real-time semantic segmentation networks.
3) We conduct comprehensive experiments on three widely used remote sensing image semantic segmentation datasets. The results show that our PyramidMamba achieves competitive accuracy compared to the state-of-the-art CNN- and Transformer-based methods.

II. RELATED WORK

*A. CNN-based Semantic Segmentation*

Semantic segmentation is a basic interpretation tool for remote sensing image understanding. In the past decade, Convolutional Neural Networks (CNNs) have taken advantage of their hierarchical structure, automatic feature learning, and end-to-end manner, dominating semantic segmentation of remote sensing images [21], [22], [23], [24]. The Fully Convolutional Network (FCN)[12] is the first end-to-end CNN-based segmentation network, marking a significant advancement in the field of semantic segmentation. However, the over-simpled fully connected decoder of the FCN often results in coarse segmentation maps.

To address this challenge, the symmetrical encoder-decoder architecture is developed [25]. The encoder progressively reduces the spatial dimensions of the image while increasing the number of channels to capture high-level semantic features, while the decoder gradually restores the spatial dimensions and strengthens detailed representation. The most famous networks are U-Net [13] and its variants [26], which effectively alleviate the coarse segmentation issue and maintain rich details of geo-objects. The results of the U-Net series, although have great improvements, still face challenges when dealing with complex remote sensing scenes. The limited local receptive field of these networks restricts their ability to capture global contextual information [27]. Thus, it is difficult for these networks to mine the crucial spatial dependencies between geo-objects and improve the global understanding ability for accurate segmentation of remote sensing images.

*B. Attention-based Global Context Modeling*

To address the limitations of traditional CNNs in semantic segmentation of remote sensing images, some studies introduced the attention mechanisms as a key technology for strengthening the global context modeling of CNNs. The DANet [15] proposed a dual attention mechanism that consists of channel-wise attention and spatial-wise attention to capture the global dependencies of both dimensions simultaneously. The CCNet [28] developed a criss-cross attention block that can capture dense global contextual information by the criss-cross feature fusion. Some other studies attempt to enlarge the receptive field by increasing the convolution kernel size [29] or merging multi-scale semantic features [30]. In particular, multi-scale feature fusion has been proven to be an effective way to enhance the performance of CNNs and obtain fine-grained segmentation results. The well-known PSPNet [18] proposed a spatial pyramid pooling module to extract and merge multi-scale semantic features and achieved great breakthroughs in the field of semantic segmentation. However, multi-scale features extracted by spatial pooling and upsampling operations exist in homogeneous semantic information, weakening the effectiveness of feature fusion. Besides, the above two schemes still demonstrate an excessive reliance on convolutional operations, not really getting rid of the local pattern.

In recent two years, Vision Transformers (ViTs) [17] that treat 2D image interpretation as 1D sequence modeling, are gradually becoming the mainstream methods for computer vision tasks, especially semantic segmentation [31]. In comparison with attention-based CNNs, ViTs adopt a pure self-attention structure, demonstrating more powerful global context modeling. Despite the outstanding capability of ViTs in global contextual information extraction, they have shortcomings in computational efficiency and local feature representation. To address the efficiency of ViTs, some studies focus on designing hierarchical structures [32] or developing efficient attention mechanisms, such as window-based attention [33] and linear attention [34]. As for improving local feature representation, the most common approach is fusing the local feature extracted by CNNs and the global feature extracted by ViTs [35]. This scheme, although effectively improves semantic information, fails to achieve more fine-grained multi-scale feature representation.

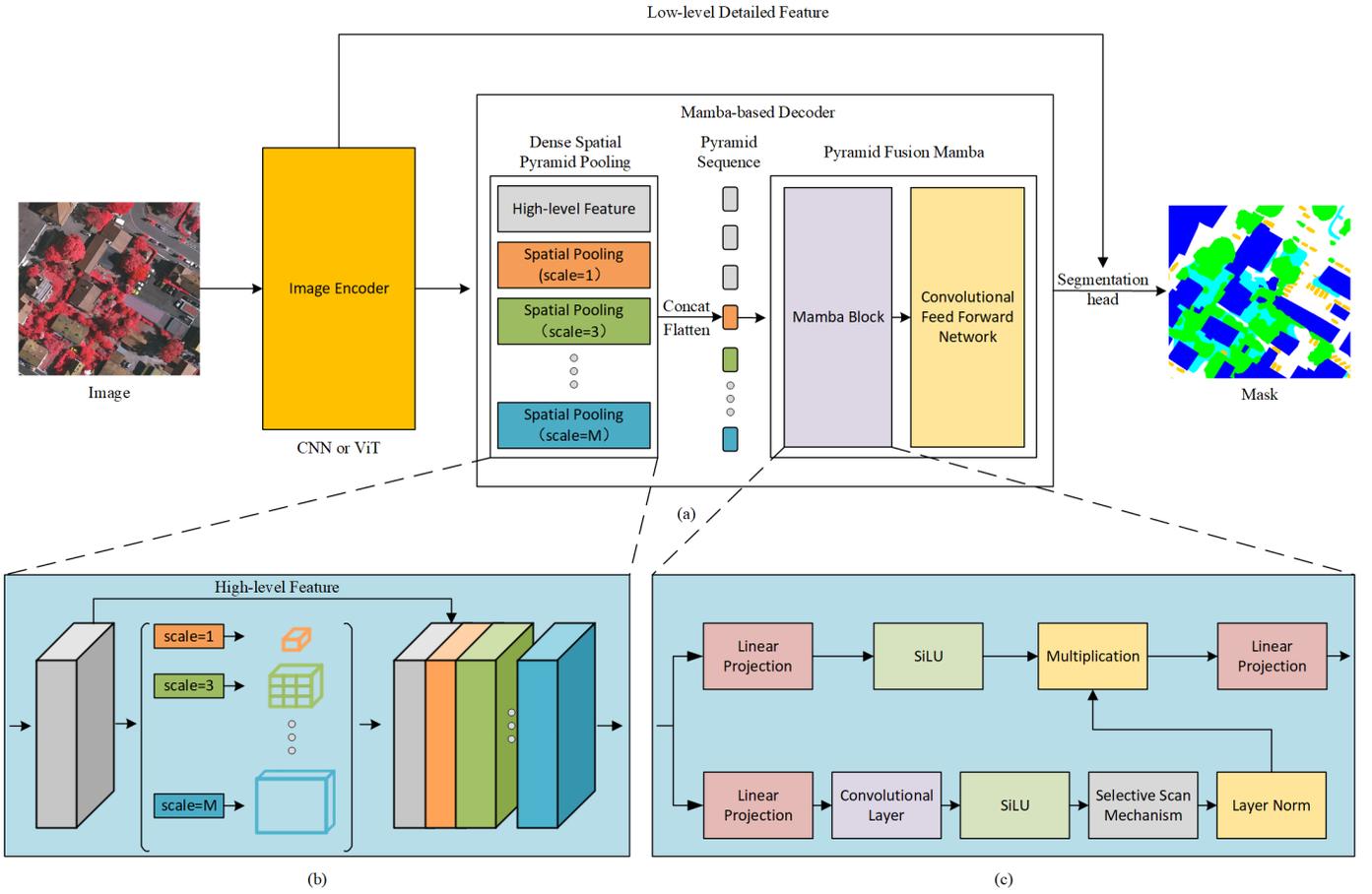

Fig. 1. An overview of our PyramidMamba. (a) Network Structure, (b) Dense Spatial Pyramid Pooling, (c) Mamba Block.

*C. Vision Mamba*

Based on the above analysis, CNN-based and ViT-based methods both have drawbacks in enlarging the receptive field. Applying multiple receptive fields with CNNs leads to homogeneous information redundancy in multi-scale feature fusion. Applying ViTs for global context modeling demonstrates a low efficiency. Recently, a novel architecture based on the selective state space model (SSM)[20], namely Mamba, opens a new way for visual understanding. Mamba uniquely employs variable parameters to represent global dependencies and leverages hardware-optimized computational strategies to balance memory efficiency and performance. Moreover, the selective scan mechanism of Mamba allows it to focus on mining the core semantics of long sequences, thereby addressing the semantic redundancy issue. Benefiting from these special characteristics, many researchers have applied Mamba successfully for computer vision tasks [36], such as image classification [37], and semantic segmentation [38]. In the field of remote sensing, Mamba has been explored for dense prediction tasks, e.g. change detection [39] and semantic segmentation [40], and obtained significant improvements compared to CNNs and ViTs.

Inspired by the selective scan mechanism and high efficiency of Mamba, we introduce it as a connector for pyramid feature fusion, thereby further enhancing multi-scale feature representation. Specifically, we integrate Mamba with a spatial pyramid module that applies more pooling scales to produce richer multi-scale semantic contents. This innovative combination achieves a win-win situation of the no-redundancy feature fusion and high efficiency.

## III. METHODOLOGY

*A. Preliminaries*

In this section, we briefly introduce the principle of the state space model (SSM), which is the basis of Mamba. More details can be found in the original paper [20].

The SSM utilizes the latent state representation $h \in \mathbb{R}^N$, subject to the input sequence $x \in \mathbb{R}^N$, in order to predict output sequence $y \in \mathbb{R}^N$. Essentially, inspired by linear time-invariant systems, the SSM maps the continuous stimulation $x \in \mathbb{R}^N$ to response $y \in \mathbb{R}^N$, which can be represented as:

$$\dot{h}(t) = Ah(t) + Bx(t)$$
$$y(t) = Ch(t) \qquad (1)$$

Where, latent state $h \in \mathbb{R}^N$ is affected by the state transition matrix $A \in \mathbb{R}^{D \times N}$, $B \in \mathbb{R}^{N \times D}$ and $C \in \mathbb{R}^{D \times N}$ are the projection matrices, where $D$ is the dimension of the input vector. However, since we generally have a discrete input (like a pixel sequence), we want to discretize the model as a zero-order hold with a time scale parameter $\Delta$, which can be defined as:

$$\overline{A} = e^{\Delta A}$$
$$\overline{B} = (\Delta A)^{-1}(e^{\Delta A} - I) \cdot \Delta B \quad (2)$$

After discretization, the output $y \in \mathbb{R}^N$ can be calculated in a convolution representation, as follows:

$$\overline{K} = (\overline{CB}, \overline{CAB}, \cdots \overline{CA}^{L-1}\overline{B})$$
$$y = x * \overline{K} \quad (3)$$

where $L$ is the length of the input sequence, $\overline{K}$ denotes the structured convolutional kernel.

Mamba further improves the state space model by introducing a selective scan mechanism, which can selectively compress the information of input tokens and output core semantics. Therefore, the model has a different matrix $B \in \mathbb{R}^{B \times L \times N}$ and $C \in \mathbb{R}^{B \times L \times N}$ for each input token, and matrix $A$ is initialized by HiPPO hardware-aware optimization techniques. Moreover, to address the challenge of dynamic variations that cannot be handled by convolution operations, a parallel scanning algorithm is employed.

The Mamba block inherits the above advantages, which is applied in our PyramidMamba. As shown in Fig.1 and 2, the selective scan mechanism first expands the pyramid sequence to four twin sequences. Then, the selective routes and S6 block [20] are applied to compress four twin sequences and extract core semantic information from each sequence. Finally, a merge operation is employed to produce the output.

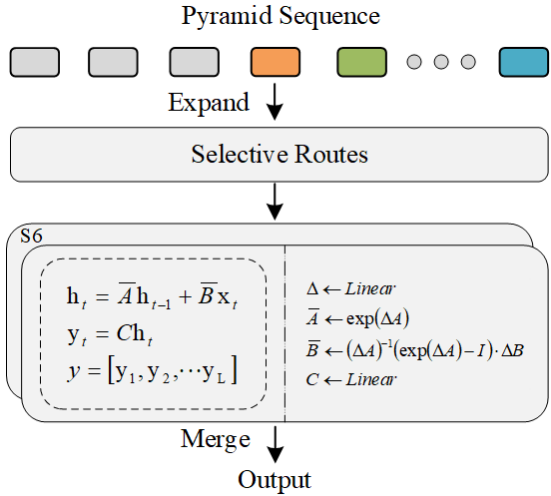

Fig. 2 The structure of the selective scan mechanism.

### B. Overview Architecture

In this section, we primarily introduce the structure of PyramidMamba, which is based on the classic encoder-decoder architecture, as shown in Fig 1. The input image is fed into the image encoder to extract the high-level feature and low-level detailed features. Then, the high-level feature is processed by the Mamba-based decoder for enhancing muti-scale semantic contents. Finally, the processed high-level feature is fused with the low-level detailed feature to strengthen the spatial details of segmentation results. The components of the Mamba-based decoder, i.e. the dense spatial pyramid pooling and pyramid fusion mamba, will be described in detail in the following sections.

### C. Image Encoder

For fine-grained semantic segmentation, it is very essential to simultaneously retain the low-level details and high-level semantic information. In our PyramidMamba, we introduce a hierarchical image encoder to extract both low-level detailed features and high-level semantic features from the input remote-sensing images. Specifically, we use the lightweight CNN (ResNet18) [41] and the window-based ViT (Swin-Base) [33] as the encoder. Consequently, there are two versions of PyramidMamba. The combination with the ResNet18 constructs a lightweight segmentation network for real-time applications. Meanwhile, the employment of the Swin-base constructs a large segmentation model to produce more precise segmentation results. Moreover, the easy switching of image encoders can illustrate the practicality of our Mamba-based decoder.

### D. Mamba-based Decoder

To address the issue of information redundancy in multi-scale feature fusion, we designed a Mamba-based decoder. Specifically, we first developed a dense spatial pyramid pooling to obtain feature maps with rich multi-scale semantic contents. Then, we employ a Mamba block which can use its selective filtering mechanism to reduce the semantic redundancy in multi-scale feature maps. Finally, a convolutional feedforward neural network is incorporated to further enhance multi-scale feature representation.

*1) Dense spatial pyramid pooling (DSPP)*: The DSPP applies different pooling scales to encode multi-scale features, as shown in Fig.1 (a) and (b). Let $X \in \mathbb{R}^{C \times N \times N}$ denotes the high-level feature map extracted by the encoder, where the DSPP can be defined as follows:

$$P_i = ConvPool_i(X), \quad i \in A_{step}^{(1,M)} \quad (4)$$

Where C and N are the channel dimension and resolution of the high-level feature map. $ConvPool_i$ represents a average pooling operation with a pooling scale $i$ and a standard convolutional layer with a kernel size of 1. $P_i \in \mathbb{R}^{c \times i \times i}$ are the pooled feature maps. The pooling scale $i$ is from the arithmetic sequence $A_{step}^{(1,M)}$, where the max value $M = N - 1$ and the margin value $step = \frac{N}{8}$.

These pooled feature maps are then upsampled by a bilinear interpolation operation to match the size of the high-level feature map:

$$U_i = BilinearUp(P_i) \quad (5)$$

where $U_i \in \mathbb{R}^{C \times N \times N}$. These upsampled feature maps are concatenated along the channel dimension, forming a multi-scale feature map:

$$X_{ms} = Concat(X, U_1, \ldots, U_M) \quad (6)$$

where $X_{ms} \in \mathbb{R}^{\left[\left(\frac{M-1}{2}+1\right)c+C\right] \times N \times N}$. However, applying bilinear interpolation operations on the pooled features from the same feature map will produce a large amount of homogeneous semantic information, resulting in the redundancy issue in the multi-scale feature.

*2) Pyramid fusion Mamba (PFM)*: Thus, we deploy the standard Mamba Block to further process the multi-scale feature, as shown in Fig.1 (a) and (c), which can use its own selective filtering mechanism to effectively characterize the core semantic across scales.

We first apply a flatten operation to generate a pyramid sequence from the multi-scale feature:

$$X_{flat} = Flatten(X_{ms}) \quad (7)$$

where $X_{flat} \in \mathbb{R}^{N^2 \times [(\frac{M-1}{2}+1)c+C]}$ denotes the pyramid sequence which is further fed into the Mamba block () for selective feature extraction. This step can be represented as:

$$X_{select} = Mamba(X_{flat}) \quad (8)$$

Finally, the selectively extracted multi-scale feature $X_{select}$ is fed into a Convolutional Feedforward Network (ConvFFN) for feature representation enhancement. The ConvFFN consists of a series of convolutional and normalization layers, followed by non-linear activation and dropout operations. The entire ConvFFN is defined as follows:

$$X_{conv} = ConvBNReLU(X_{select}) \quad (9)$$

Where the $ConvBNReLU$ represents a one-dimension $1 \times 1$ convolutional layer followed by the batch normalization and ReLU activation. Then, two fully connected one-dimension convolutional layers with a GELU activation function and a dropout operation are applied for regularization:

$$X_{fc1} = Conv(X_{conv}) \quad (10)$$

$$X_{act} = GELU(X_{fc1}) \quad (11)$$

$$X_{drop} = Dropout(X_{conv}) \quad (12)$$

$$X_{fc2} = Conv(X_{drop}) \quad (13)$$

$$X_{out} = Dropout(X_{fc2}) \quad (14)$$

By following such a design, the Mamba-based decoder effectively aggregates multi-scale features, reduces information redundancy, and enhances the multi-scale feature representation for fine semantic segmentation.

*E. Loss Function*

To better deal with the common class-imbalance issue in semantic segmentation datasets, we adopt a joint loss to train our PyramidMamba. The joint loss function $L$ can be defined as:

$$L = L_{ce} + L_{dice} \quad (15)$$

$$L_{ce} = -\frac{1}{N} \sum_{n=1}^{N} \sum_{k=1}^{K} y_k^{(n)} log \hat{y}_k^{(n)} \quad (16)$$

$$L_{dice} = 1 - \frac{2}{N} \sum_{n=1}^{N} \sum_{k=1}^{K} \frac{y_k^{(n)} \hat{y}_k^{(n)}}{\hat{y}_k^{(n)} + y_k^{(n)}} \quad (17)$$

where $N$ and $K$ denote the number of pixels and the number of categories, respectively. $L_{ce}$ is the cross-entropy loss. $L_{dice}$ is the dice loss. $y_k^{(n)}$ represents the true label, and $\hat{y}_k^{(n)}$ denotes the confidence of the pixel $n$ belonging to the category $k$.

## IV. EXPERIMENTAL SETTINGS AND DATASETS

*A. Datasets*

To evaluate the performance of the proposed PyramidMamba, three publicly available remote sensing semantic segmentation datasets are used for conducting experiments, including the OpenEarthMap dataset [42], the ISPRS Vaihingen dataset, and the ISPRS Potsdam dataset. The following are the details of the datasets.

*1) OpenEarthMap*: The OpenEarthMap dataset is a large-scale high-resolution land cover mapping dataset, which consists of 5000 images and contains eight land cover classes (bareland, rangeland, developed space, road, tree, water, agriculture land, building). The image spatial resolution is ranging from 0.25m to 0.5m. The spatial distribution covers 97 regions from 44 countries across six continents. Semantic segmentation of the OpenEarthMap dataset is very challenging due to its wide spatial variation, complex geo-objects, and scenes. In the OpenearthMap dataset, the remote sensing images of each region were randomly divided into training, validation, and test sets, which yielded 3000, 500, and 1500 images, respectively. In our experiments, we use the validation set for quantitative comparisons since the test set is not open public. The input images were uniformly resized to $1024 \times 1024$ px patches, and data augmentation strategies like horizontal and vertical flips were used in the training and testing phase.

*2) Vaihingen*: The Vaihingen dataset consists of 33 fine-resolution image tiles at an average size of $2494 \times 2064$ pixels. Each image tile has three multispectral bands (near-infrared, red, green) as well as a digital surface model (DSM) and normalized digital surface model (NDSM) with a 9 cm ground sampling distance (GSD). The dataset involves five foreground classes (impervious surface, building, low vegetation, tree, car) and one background class (clutter). In our experiments, only the image tiles were used. The image tiles were cropped into $1024 \times 1024$ px patches. Data augmentation strategies including horizontal and vertical flips, random scaling and cropping, and random mosaic were used for training models.

*3) Potsdam*: The Potsdam dataset consists of 38 ultra-high resolution aerial images (Ground sample distance 5 cm) at a size of $6000 \times 6000$ pixels and involves 6 geo-object categories (impervious surface, low vegetation, tree, car, building, and clutter), four spectral bands (red, green, blue, and near-infrared), as well as the DSM and NDSM. In our experiments, we followed the official partition for training and testing, and only three bands (red, green, and blue) were used. The original image tiles were cropped into $1024 \times 1024$ px patches as the input, and we applied random flip and random mosaic as the data augmentation.

*B. Evaluation Metrics*

We use the overall accuracy (OA), mean intersection over union (mIoU), F1 score, precision, and recall to evaluate the performance of models, which can be defined as follows:

$$OA = \frac{\sum_{k=1}^{N} TP_k}{\sum_{k=1}^{N} TP_k + FP_k + TN_k + FN_k}, \quad (18)$$

$$mIoU = \frac{1}{N} \sum_{k=1}^{N} \frac{TP_k}{TP_k + FP_k + FN_k}, \quad (19)$$

$$precision = \frac{1}{N}\sum_{k=1}^{N}\frac{TP_k}{TP_k + FP_k}, \tag{20}$$

$$recall = \frac{1}{N}\sum_{k=1}^{N}\frac{TP_k}{TP_k + FN_k}, \tag{21}$$

$$F1 = 2 \times \frac{precision \times recall}{precision + recall}, \tag{22}$$

where $TP_k$, $FP_k$, $TN_k$, and $FN_k$ indicate the true positive, false positive, true negative, and false negatives, respectively, for the specific object indexed as class $k$. OA is computed for all categories including the background pixels.

### C. Experimental Setting

All deep models in the experiments were implemented with the PyTorch framework on a single NVIDIA GTX 4090 GPU. The AdamW optimizer was employed to train deep models. The Poly learning rate adjusting strategy was used and the power parameter was set to 0.9. The base learning rate was set to 6e-4 while the learning rate for the image encoder was specially set to 6e-5. The batch size and weight decay were set to 2 and 0.01, respectively. The total training epoch was set to 45 and the warmup training strategy was applied for the first 5 epochs. An early stopping strategy was used to prevent over-fitting. In the testing phase, we applied data augmentation technologies like horizontal and vertical flipping as well as multiple scales, which is also known as test-time augmentation (TTA).

### D. Benchmark Methods

To verify the effectiveness of the proposed method, we selected a set of state-of-the-art segmentation methods for comprehensive comparisons, including: 1) Real-time semantic segmentation networks: BiSeNet [43], ShelfNet [44], SwiftNet [45], ABCNet [46], and UNetFormer [47], 2) CNN-based semantic segmentation networks: U-Net [13], PSPNet [18], DeepLabV3+ [30], DANet [15], UFMG-4 [48], ResUNet-a [22], MANet [27], LANet [21], DDCM-Net [49], EaNet [50], 3) Transformer-based semantic segmentation networks: SegFormer [51], Segmenter [31], SwinUperNet [33], BoTNet [52], DC-Swin [53], SwinB-CNN [54], CG-Swin [55], Mask2Former [56], 4) Mamba-based semantic segmentation networks: RS³Mamba [57], 5) Vision language models for semantic segmentation: CLIPSeg [58], 6) Recent remote sensing image segmentation networks: FTransUNet [59], SAPNet [60] and MMT [61].

## V. EXPERIMENTAL RESULTS AND ANALYSIS

### A. Ablation Study

To verify the effectiveness of the proposed modules, we conducted ablation experiments on the ISPRS Vaihingen dataset. To ensure the fairness of ablation experiments, we did not apply any test-time augmentation in the testing phase and the image encoder was uniformly set to ResNet18.

*1) Network variants*: As shown in Table I, the Baseline consists of the image encoder and upsampling operations. The Baseline+DSPP indicates the combination of Baseline and the dense spatial pyramid pooling, and the Baseline+DSPP+PFM represents the entire network without the low-level detailed feature.

TABLE I
THE ABLATION EXPERIMENTAL RESULTS OF EACH COMPONENT ON THE VAIHINGEN DATASET.

| Method | mIoU | F1 |
|---|---|---|
| Baseline | 74.6 | 84.5 |
| Baseline+DSPP | 78.1 | 87.1 |
| Baseline+DSPP+PFM | 79.2 | 87.9 |

*2) The effectiveness of each component*: In the proposed PyramidMamba, the DSPP encodes rich multi-scale semantic information with a simple concat operation for fine-grained segmentation. As listed in Table I, the deployment of the DSPP provides an increase of 3.5% in mIoU, which can illustrate its effectiveness in multi-scale feature representation. Besides, the utilization of PFM can further improve the mIoU by 1.1%. This result not only demonstrates the effectiveness of PFM but also indicates the significant advantages of Mamba for multi-scale feature fusion.

*3) The superiority of the dense spatial pyramid pooling (DSPP)*: As shown in Table II, we compare the proposed DSPP with the standard spatial pyramid pooling (SPP) module in PSPNet. The Baseline was selected as the basic network. The results show that our DSPP yields an improvement of 1.4% in mIoU and 0.9% in F1 score compared to the SPP, which can illustrate the superiority of dense pooling in strengthening multi-scale representation.

TABLE II
THE ABLATION EXPERIMENTAL RESULTS OF THE DENSE POOLING ON THE VAIHINGEN DATASET.

| Method | mIoU | F1 |
|---|---|---|
| Baseline+SPP | 76.7 | 86.2 |
| Baseline+DSPP | 78.1 | 87.1 |

*4) The effectiveness of aggregating the low-level detailed feature (LDF)*: Introducing a spatial detailed feature is an effective way of optimizing semantic segmentation results. The low-level features of hierarchical deep networks involve rich spatial details due to their higher resolutions. To demonstrate the contribution of the low-level detailed feature to accuracy, we remove it for ablation experiments. As listed in Table III, the employment of the low-level detailed feature increases the mIoU metric and the F1 score by 2.7% and 2.1%, respectively, demonstrating its effectiveness and necessity.

TABLE III
THE ABLATION EXPERIMENTAL RESULTS OF THE LOW-LEVEL DETAILED FEATURE ON THE VAIHINGEN DATASET.

| Method | mIoU | F1 |
|---|---|---|
| PyramidMamba w/o LDF | 79.2 | 87.9 |
| PyramidMamba w/ LDF | 81.9 | 89.8 |

### B. Comparisons of State-of-the-art Real-time Semantic Segmentation Methods

The combination of the DSPP and PFM not only has advantages in terms of accuracy but also has great potential in constructing real-time segmentation networks due to its high efficiency. To better demonstrate this point, we conduct comprehensive experiments in comparison with other advanced real-time semantic segmentation. Notably, the test-time augmentation was used for a fair comparison in this section. As shown in Table IV, the speed of the network (FPS) is measured by two 1024×1024 image patches on a single NVIDIA GTX 4090 GPU. The results reveal that the proposed PyramidMamba has advantages in accuracy while keeping a competitive speed compared to other advanced real-time segmentation networks. Specifically, our PyramidMamba yielded an increase of 0.4% mIoU in comparison with the recent real-time ViT (UNetFormer) and outperformed other real-time CNNs by at least 1.8% mIoU. These results not only demonstrated the superiority of our PyramidMamba but also illustrated the great potential of Mamba-based methods in building real-time deep networks.

TABLE IV
QUANTITATIVE COMPARISON WITH STATE-OF-THE-ART REAL-TIME SEMANTIC SEGMENTATION METHODS ON THE VAIHINGEN DATASET.

| Method | Backbone | Speed (FPS) | mIoU |
|---|---|---|---|
| BiSeNet [43] | ResNet18 | 142.2 | 75.8 |
| ShelfNet [44] | ResNet18 | 157.7 | 78.3 |
| SwiftNet [45] | ResNet18 | 149.9 | 79.6 |
| ABCNet [46] | ResNet18 | 108.0 | 81.3 |
| Segmenter [31] | ViT-Tiny | 23.3 | 73.6 |
| UNetFormer [47] | ResNet18 | 132.1 | 82.7 |
| PyramidMamba (Ours) | ResNet18 | 73.8 | 83.1 |

## C. Quantitative Comparisons with State-of-the-art Semantic Segmentation Methods

To further verify the effectiveness of the proposed method, we compare it with state-of-the-art methods on three publicly available datasets, i.e. the OpenEarthMap dataset, ISPRS Vaihingen, and Potsdam dataset. Moreover, to ensure a fair comparison and demonstrate the applicability of our Mamba-based decoder, the image encoder is set to a widely-used ViT, i.e. Swin-Base.

*1) OpenEarthMap*: The OpenEarthMap dataset includes many complex scenes and confusing geo-objects. Thus, it is very challenging to achieve a high accuracy on this dataset. As shown in Table V, our PyramidMamba yields a 70.8% mIoU, surpassing the CNN-based method MANet and the ViT-based method SegFormer by 6.8% and 4.8% in mIoU, respectively. Notably, our method also achieves the highest accuracy on special categories, such as IoU-Road (64.9%), IoU-Building (79.6%), and IoU-Developed (57.9%). The significant improvements on these multi-scale geo-objects can demonstrate the effectiveness and superiority of our PyramidMamba in multi-scale feature representation. The visualization results can further witness this point. As shown in Fig. 2, in comparison with the UNet and UNetFormer, the proposed method not only can segment buildings with fine shapes (first row) but also maintains the continuity of the road (second row). Besides, as for the confusing developed land, our PyramidMamba also has significant advantages.

TABLE V
QUANTITATIVE COMPARISONS WITH STATE-OF-THE-ART METHODS ON THE OPENEARTHMAP DATASET. THE BOLD DENOTES THE BEST VALUES. THE UNDERLINE DENOTES THE SECOND-BEST VALUES.

| Method | Background | Bareland | Rangeland | Developed | Road | Tree | Water | Agriculture | Building | mIoU |
|---|---|---|---|---|---|---|---|---|---|---|
| UNet [13] | 82.0 | 23.6 | 52.3 | 49.8 | 57.3 | 67.2 | 67.1 | 71.7 | 73.0 | 60.4 |
| DANet [15] | 81.1 | 30.9 | 49.8 | 47.5 | 57.4 | 63.3 | 67.6 | 73.2 | 70.2 | 60.1 |
| CLIPSeg [58] | 95.5 | 34.5 | 46.1 | 40.8 | 45.2 | 60.0 | 69.1 | 73.7 | 62.2 | 58.6 |
| BoTNet [52] | 81.9 | 30.3 | 50.9 | 50.2 | 57.9 | 67.8 | 65.5 | 75.8 | 72.9 | 61.5 |
| MANet [27] | 91.0 | 41.2 | 50.8 | 50.8 | 50.8 | 60.0 | 69.3 | 70.1 | 75.4 | 64.0 |
| SegFormer [51] | 97.2 | 41.0 | 56.4 | 53.2 | 58.7 | 70.9 | 77.7 | 76.7 | 76.2 | 66.0 |
| DC-Swin [53] | 96.4 | 42.5 | 55.6 | 53.2 | 58.1 | 69.9 | 77.8 | 76.0 | 75.7 | 67.2 |
| UNetFormer [47] | 97.2 | 42.8 | 56.2 | 53.5 | 60.9 | 70.2 | 77.4 | 76.6 | 76.9 | 68.0 |
| RS³Mamba [57] | 96.3 | 39.9 | 51.0 | 48.7 | 56.9 | 66.8 | 74.4 | 75.0 | 71.4 | 64.5 |
| PyramidMamba(Ours) | **97.4** | **45.0** | **59.4** | **57.9** | **64.9** | **72.1** | **81.3** | **79.4** | **79.6** | **70.8** |

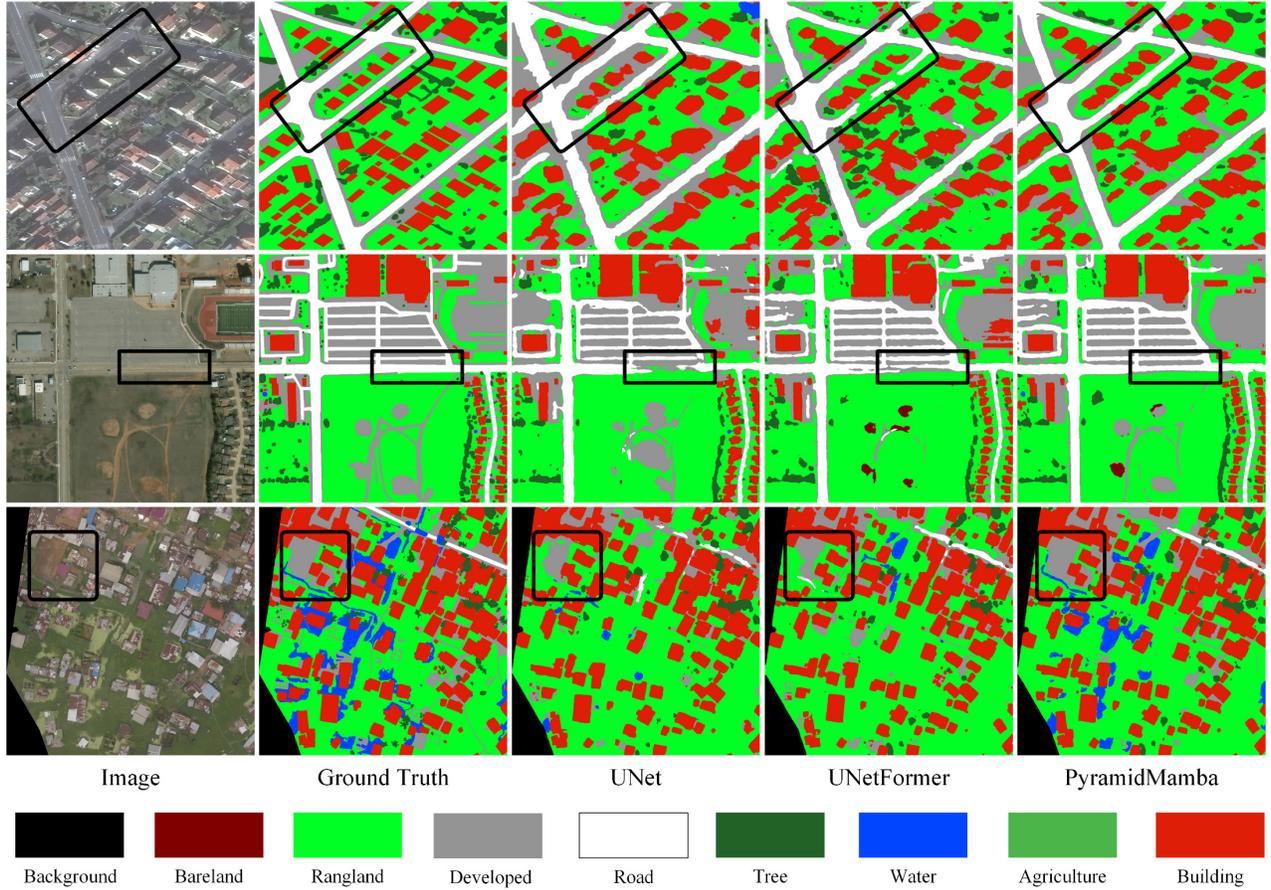

Fig. 2 Visual comparisons on the OpenEarthMap dataset.

*2) Vaihingen*: The ISPRS Vaihingen dataset is a widely used dataset for verifying the effectiveness of remote sensing image semantic segmentation methods. Hundreds of deep models have been developed and achieved high scores on this dataset. Thus, it is difficult to obtain further accuracy breakthroughs. However, as shown in Table VI, our PyramidMamba yields the best mIoU (84.8%) and overall accuracy (93.7%). In particular, our PyramidMamba outperforms the recent remote sensing image segmentation methods, SAPNet and MMT, by 4.0% in OA and 0.7% in mIoU, which can demonstrate the advancement and superiority of our method. Moreover, as for the RS$^3$Mamba that adopts the same Mamba-based architecture, our PyramidMamba yields an increase of 1.9% in mIoU. Meanwhile, the visualization results can further witness the advantages of our method. As shown in Fig.3, benefiting from the fine-grained multi-scale feature fusion, our PyramidMamba can ensure the integrity of building segmentation despite the surface of the buildings being very complex. As for the tiny geo-object car, our method can also keep a good segmentation shape.

TABLE VI
QUANTITATIVE COMPARISONS WITH STATE-OF-THE-ART METHODS ON THE VAIHINGEN DATASET. THE BOLD DENOTES THE BEST VALUES. THE UNDERLINE DENOTES THE SECOND-BEST VALUES.

| Method | Imp. surf. | Building | Low. veg. | Tree | Car | Mean F1 | OA | mIoU |
|---|---|---|---|---|---|---|---|---|
| PSPNet [18] | 92.7 | 95.4 | 84.5 | 89.9 | 88.6 | 90.2 | 90.8 | 82.5 |
| DeepLabV3+ [30] | 92.3 | 95.1 | 84.2 | 89.5 | 86.4 | 89.5 | 90.5 | 81.4 |
| EaNet [50] | 93.4 | <u>96.2</u> | **85.6** | 90.5 | 88.3 | 90.8 | 91.2 | - |
| UFMG_4 [48] | 91.1 | 94.5 | 82.9 | 88.0 | 81.3 | 87.7 | 89.4 | - |
| MANet [27] | 93.0 | 95.4 | 84.6 | 89.9 | 88.9 | 90.4 | 90.9 | 82.7 |
| SwinUperNet [33] | 92.8 | 95.6 | 85.1 | 90.6 | 85.1 | 89.8 | 91.0 | 81.8 |
| Mask2Former [56] | <u>96.9</u> | 92.9 | 86.2 | 90.3 | 87.8 | 90.3 | 90.3 | 83.0 |
| RS$^3$Mamba [57] | 96.6 | 95.5 | 83.8 | 89.5 | 86.5 | 90.4 | <u>92.9</u> | 82.9 |
| SAPNet [60] | 94.2 | <u>96.2</u> | **85.6** | <u>91.6</u> | **91.7** | **91.8** | 89.7 | - |
| FTransUNet [59] | 93.0 | **98.2** | 81.4 | **91.9** | <u>91.2</u> | 91.2 | - | <u>84.2</u> |
| PyramidMamba (Ours) | **97.0** | 96.1 | <u>85.5</u> | 90.3 | 89.2 | <u>91.6</u> | **93.7** | **84.8** |

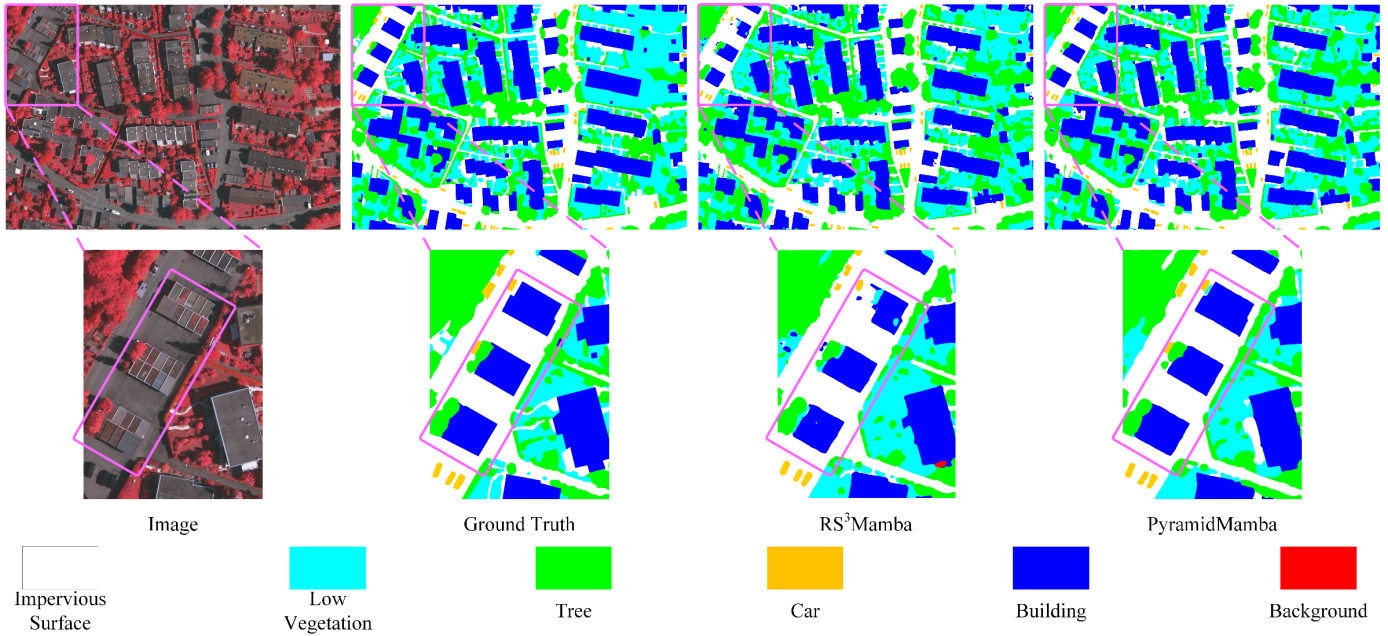

Fig. 3 Visual comparisons on the Vaihingen dataset.

*3) Potsdam*: The ISPRS Potsdam dataset is also a widely used dataset for remote sensing image semantic segmentation. On this dataset, our PyramidMamba yields the state-of-the-art mIoU (88.0%), mean F1 score (93.5%), and IoU-Car (96.9%), outperforming CNN-based methods by at least 1.1% in mIoU and Transformer-based methods more than 0.3% in mean F1 score. These results further demonstrate the effectiveness and superiority of our PyramidMamba. We also conduct a visual comparison with the Transformer-based method CG-Swin. As shown in Fig.4, our PyramidMamba has significant advantages in detecting narrow roads.

TABLE VII
QUANTITATIVE COMPARISONS WITH STATE-OF-THE-ART METHODS ON THE POTSDAM DATASET. THE BOLD DENOTES THE BEST VALUES. THE UNDERLINE DENOTES THE SECOND-BEST VALUES.

| Method | Imp. surf. | Building | Low. veg. | Tree | Car | Mean F1 | OA | mIoU |
|---|---|---|---|---|---|---|---|---|
| PSPNet [18] | 93.3 | 96.9 | 87.7 | 88.5 | 95.4 | 92.4 | 91.0 | 84.8 |
| DeepLabV3+ [30] | 92.9 | 95.8 | 87.6 | 88.1 | 96.0 | 92.1 | 90.8 | 84.3 |
| DDCM-Net [49] | 92.9 | 96.9 | 87.7 | 89.4 | 94.9 | 92.3 | 90.8 | 86.0 |
| LANet [21] | 93.0 | 97.1 | 87.3 | 88.0 | 94.1 | 91.9 | 90.8 | - |
| ResUNet-a [22] | 93.4 | 96.9 | 88.3 | 89.3 | 96.4 | 92.9 | 91.3 | 86.9 |
| SwinB-CNN [54] | 93.6 | 96.7 | 88.0 | 88.0 | 96.3 | 92.5 | 91.0 | - |
| CG-Swin [55] | 94.0 | <u>97.4</u> | 88.5 | 89.7 | <u>96.6</u> | <u>93.2</u> | 91.9 | 87.6 |
| Mask2Former [56] | <u>97.9</u> | 96.9 | 88.4 | 90.6 | 84.5 | 92.5 | <u>92.5</u> | 86.6 |
| SAPNet [60] | 94.4 | **97.7** | 87.8 | 89.6 | 96.0 | 93.1 | 91.8 | - |
| MMT [61] | **99.8** | 97.3 | **89.6** | 90.6 | 93.6 | <u>93.2</u> | **93.4** | <u>87.7</u> |
| PyramidMamba (Ours) | 94.7 | 97.2 | <u>88.6</u> | <u>89.9</u> | **96.9** | **93.5** | 92.3 | **88.0** |

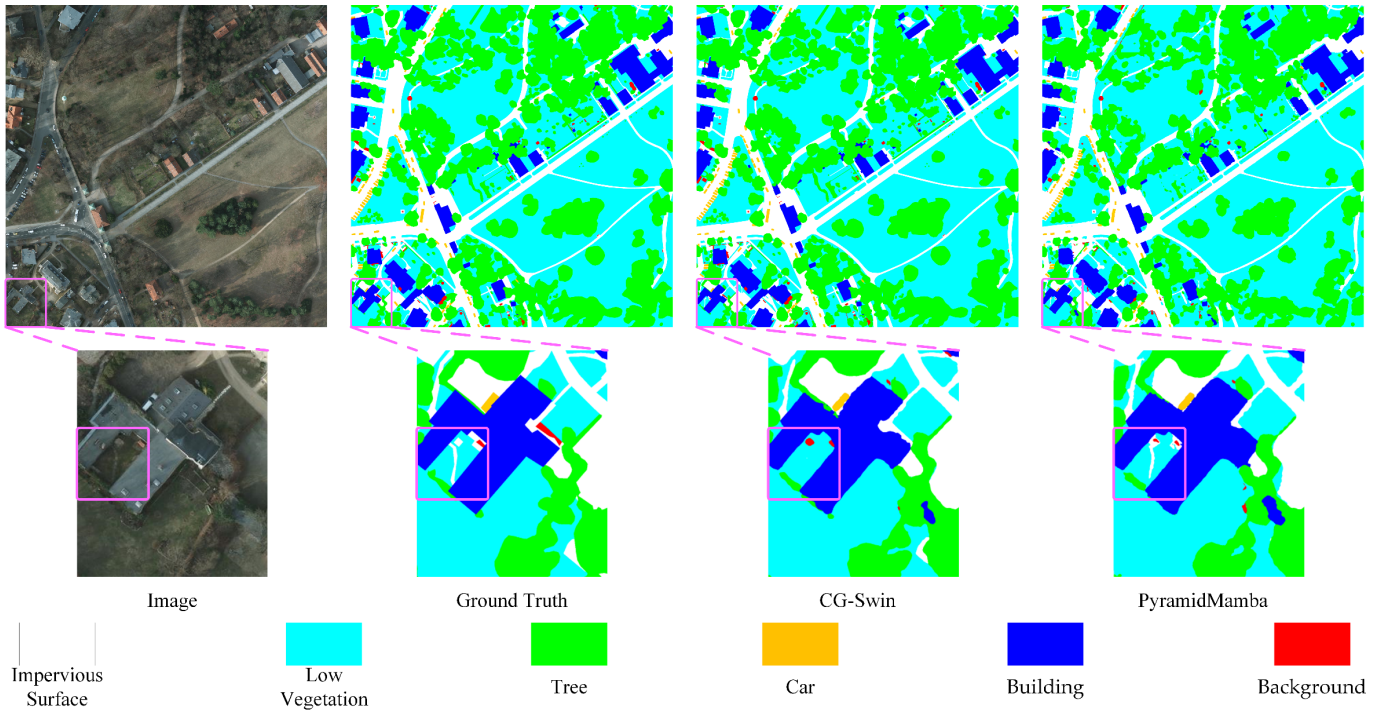

Fig. 4 Visual comparisons on the Potsdam dataset.

## VI. Conclusion

In this paper, we proposed a novel Mamba-based decoder for semantic segmentation of remote sensing images, namely PyramidMamba. To address the semantic redundancy issue in multi-scale feature fusion, we introduce a standard Mamba block into the decoder, taking advantage of its selective scanning mechanism to enhance multi-scale feature representation. Furthermore, we proposed a dense spatial pyramid pooling to achieve the fine-grained pyramid features. Benefiting from the above, our PyramidMamba demonstrated superior in comparison with state-of-the-art methods on three publicly available and widely-used remote sensing image segmentation datasets. Meanwhile, ablation studies also illustrate the effectiveness of each component in the proposed decoder and reveal its great potential in constructing real-time semantic segmentation networks. In the future, we will continue to explore the potential of Mamba-based structure in multimodal learning and foundational models.